\documentclass[letterpaper, 10pt, conference]{ieeeconf}
\IEEEoverridecommandlockouts 
\usepackage[export]{adjustbox}
\usepackage{cite}
\usepackage{amsmath}
\usepackage{amssymb}
\usepackage{amsfonts}
\usepackage{bm}
\usepackage{graphicx}
\usepackage[caption=false]{subfig}

\usepackage{algorithm}
\usepackage{algorithmic}

\usepackage{glossaries}

\usepackage[binary-units=true,separate-uncertainty=true,multi-part-units=single,per-mode=symbol]{siunitx}

\usepackage{tabularx}
\usepackage{multirow}

\newcolumntype{C}{>{\centering\arraybackslash}X} 

\usepackage{paralist}

\renewcommand{\baselinestretch}{0.99}

\makeatletter
\let\NAT@parse\undefined
\makeatother
\usepackage[bookmarks=true,hidelinks]{hyperref}  

\newacronym{SDF}{SDF}{Signed Distance Field}
\newacronym{ESDF}{ESDF}{Euclidean Signed Distance Field}
\newacronym{USDF}{USDF}{Unsigned Distance Field}
\newacronym{TSDF}{TSDF}{Truncated Signed Distance Field}
\newacronym{EDT}{EDT}{Euclidean Distance Transform}
\newacronym{MAV}{MAV}{Micro Aerial Vehicle}
\newacronym{HSR}{HSR}{Toyota Human Support Robot}

\hypersetup{
    pdfauthor   = {Mark Nicholas Finean},%
    pdftitle    = {Simultaneous Scene Reconstruction and Whole-Body Motion Planning for Safe Operation in Dynamic Environments},%
    pdfsubject  = {Robotics},%
    pdfkeywords = {Robots, Trajectory Optimization, Signed Distance Fields, Collision Avoidance, Dynamic Environments, Real-time, Moving Obstacles}%
}

\DeclareMathAlphabet\mathbfcal{OMS}{cmsy}{b}{n}  

\title{\LARGE\bf
  Simultaneous Scene Reconstruction and Whole-Body Motion Planning for Safe Operation in Dynamic Environments
}

\author{
    Mark Nicholas Finean$^1$, Wolfgang Merkt$^1$, and Ioannis Havoutis$^1$%
    \thanks{$^1$ Oxford Robotics Institute, University of Oxford}%
    \thanks{This work was supported by the UK Engineering and Physical Sciences Research Council (EPSRC), award reference 1904254, for the University of Oxford Centre for Doctoral Training, Autonomous Intelligent Machines and Systems (AIMS). This work was partly supported by the UKRI/EPSRC grants [EP/S002383/1], [EP/R026084/1] and [EP/R026173/1] and the EU H2020 Project MEMMO(780684). This work was part of the Human-Machine Collaboration Programme, supported by a gift from Amazon Web Services.
Email: \texttt{mfinean@robots.ox.ac.uk}}%
}

\begin{document}
\bstctlcite{IEEEexample:BSTcontrol} 

\maketitle
\thispagestyle{empty}
\pagestyle{empty}

\begin{abstract}
%
Recent work has demonstrated real-time mapping and reconstruction from dense perception, while motion planning based on distance fields has been shown to achieve fast, collision-free motion synthesis with good convergence properties. However, demonstration of a fully integrated system that can safely re-plan in unknown environments, in the presence of static and dynamic obstacles, has remained an open challenge.
In this work, we first study the impact that signed and unsigned distance fields have on optimisation convergence, and the resultant error cost in trajectory optimisation problems in 2D path planning, arm manipulator motion planning, and whole-body loco-manipulation planning. We further analyse the performance of three state-of-the-art approaches to generating distance fields (Voxblox, Fiesta, and GPU-Voxels) for use in real-time environment reconstruction. Finally, we use our findings to construct a practical hybrid mapping and motion planning system which uses GPU-Voxels and GPMP2 to perform receding-horizon whole-body motion planning that can smoothly avoid moving obstacles in 3D space using live sensor data. Our results are validated in simulation and on a real-world \acrfull{HSR}.
\end{abstract}

\section{Introduction}
In recent years, we have seen tremendous advances across many fields of robotics, from hardware to vision and planning. As the capabilities of robots has increased, the question is now ``when will we see large scale integration into our daily lives?". A key concern that needs to be solved before robots become commonplace is that of safety; we require robots to be reliable and interact safely with their surroundings. Key hereto is the ability to recognise and reason about static and dynamic obstacles in real-time to prevent collision and injury to people, the environment, and the robots themselves.

There is significant research in motion planning that focuses on or assumes a static environment, however, this assumption breaks down in the real-world where our surroundings are often dynamic. For robots to become more widely used, such as in household environments, they must be able to perform motion planning and collision avoidance in the presence of moving obstacles. 
Considerable research in the mapping and scene reconstruction communities has achieved the ability to reconstruct environments in great detail in real-time based on voxel grids \cite{Niesner2013,GPUVoxels}, \acrfullpl{TSDF} \cite{KinectFusion,Whelan2012}, surfels \cite{Whelan2015,Scona2018}, and octrees representations \cite{Steinbrucker2014} using CPU or GPU computation.
For instance, this capability has been used in path planning for \acrfullpl{MAV} \cite{Fiesta, voxblox, GPUVoxelsMobile, Usenko2017}. At the same time, work in the motion planning community has proposed solutions for fast planning with real-world sensed data, enabling the generation of trajectories that avoid collisions with static and dynamic obstacles in discrete-time \cite{chomp,gpmp2,ITOMP,Ratliff2015} and continuous-time \cite{trajopt,merkt2019continuous}. However, there has been little work in combining the two to provide a usable integrated framework for real-time re-planning in dynamic environments, providing motivation for the research presented here.

\begin{figure}[t]
\includegraphics[width=\columnwidth]{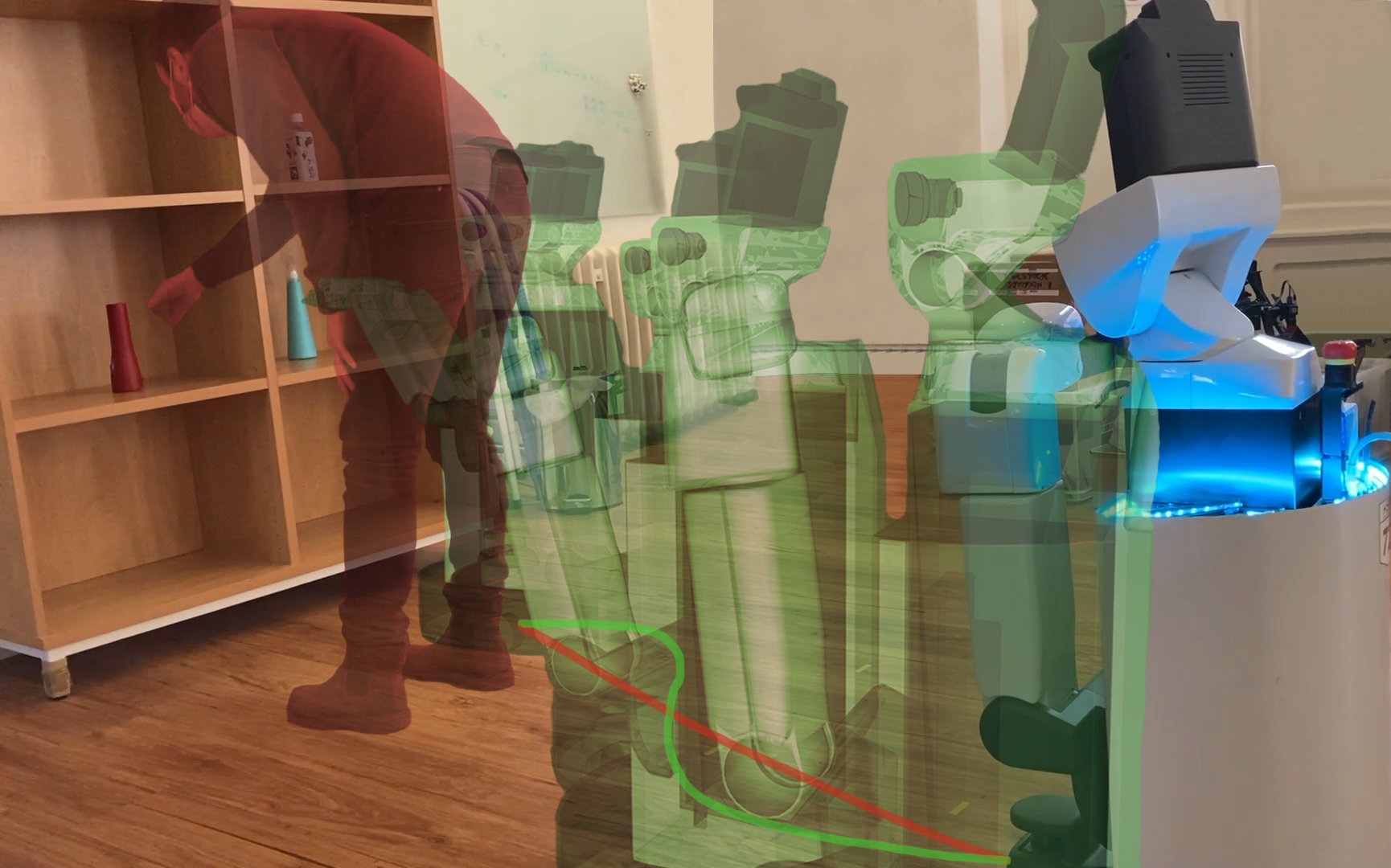}
\caption{Real-time re-planning and avoidance of moving obstacles using the \acrshort{HSR}. The robot was tasked with picking up a bottle from a shelf. During execution, a human demonstrator walked into the planned trajectory of the robot. Using our integrated motion planning and real-time mapping system, the robot successfully integrated the dynamic obstacle without any prior in real-time and re-planned around the newly observed obstacle in a receding-horizon fashion.}
\label{fig:hardware_timelapse}
\end{figure} 

In this work, we provide the first comparison of the impact that \acrfullpl{SDF} and \acrfullpl{USDF} have on trajectory optimisation convergence and the error cost for navigation planning, motion planning, and whole-body loco-manipulation. We explore a selection of state-of-the-art mapping and reconstruction packages to determine the best performing method for integration with a whole-body motion planner. After discussing our motivation for motion planner selection, we describe its integration with the reconstruction pipeline. 

Overall, we present a full framework to concurrently map the environment in 3D and perform fast re-planning online in a dynamic environment with a whole-body mobile manipulator (\acrshort{HSR}).
The key contributions of this paper are:
\begin{compactenum}
\item Integration of trajectory optimisation-based whole-body motion planning with fast GPU-based distance field reconstruction.
\item The first demonstration of real-time reactive collision avoidance using whole-body trajectory optimisation and live sensor data in unknown 3D environments for loco-manipulation on articulated systems (8 DoF).
\item Exploration of incrementally built \acrfullpl{EDT} for the motion planning of whole-body and articulated systems.
\item A comparison of the impact that \acrshortpl{USDF} and \acrshortpl{SDF} have on trajectory optimisation convergence and optimality.
\end{compactenum}

\section{Related Work}
We separate the relevant literature into three categories; we first explore the state-of-the-art in motion planning in dynamic 3D environments and make the case for using a trajectory optimisation approach (GPMP2). We then explore the options for online mapping and reconstruction of environment distance fields. Lastly, we discuss the prior work in exploring integrated systems to provide collision avoidance in real-world environments with moving obstacles.

\subsection{Motion Planning}
Motion planning is a well-studied problem with many possible approaches. Sampling-based algorithms and graph search methods \cite{LaValle1998, Kavraki1996, lavalle_2006} are among the most common techniques to apply. They are mathematically appealing because we can select for desirable characteristics such as completeness, however they often require post-processing to address smoothness of the motions. While there are optimal planners which aim to address this, they suffer from the curse of dimensionality for robots with many degrees of freedom \cite{Karaman2011}. A key consideration in this work was a motion planner's ability to plan in dynamic environments, placing emphasis on fast planning times and ideally the ability to re-plan smoothly in response to changes in the environment. Trajectory optimisation-based methods are well suited to these criteria.

Trajectory optimisation algorithms, such as CHOMP, STOMP and TrajOpt, operate by minimising an objective function to solve for a feasible and optimal trajectory \cite{chomp, Kalakrishnan2011, trajopt}. In the cases of CHOMP and STOMP, the trajectory is finely discretised to generate trajectories that avoid obstacles and maintain smoothness; however, fine discretisation is computationally expensive. TrajOpt and our prior work \cite{merkt2019continuous} represent the trajectory using fewer states by introducing continuous-time collision checking. Yet, to achieve smoothness, a fine discretisation may still be required in practice.

Another approach to reduce computational requirements is to use kernel embeddings to represent the trajectory. GPMP2 \cite{gpmp2}, for instance, leverages Gaussian Process Interpolation to achieve a continuous trajectory representation using a small number of discrete points. Similar to AICO \cite{AICO}, GPMP2 treats the motion planning problem as probabilistic inference on a factor graph. GPMP2 is built upon the GTSAM framework \cite{Dellaert2012} which uses factor graphs and Bayes networks as the underlying objects to frame an optimisation problem for the most probable configuration or plan. Factor graphs are popular in the Simultaneous Localisation and Mapping (SLAM) community and working in this paradigm offers the flexibility to use efficient tools, such as incremental Smoothing and Mapping (iSAM) to perform fast incremental inference for re-planning \cite{Kaess2008, Kaess2012}. Mukadam et. al. show iSAM-based re-planning (iGPMP2) to be an order of magnitude faster than planning from scratch \cite{gpmp2}. The superior planning speed and ability to quickly re-plan makes the GPMP2 factor graph formulation an appealing candidate for use in dynamic environments.

\subsection{Collision Avoidance}
Some approaches in the literature view obstacles from a more theoretical viewpoint and assume prior knowledge of their shape, size and position. Ratliff et. al. model obstacle constraints as analytical inequality constraints with a margin in RieMo \cite{Ratliff2015}. Merkt et. al. \cite{merkt2018leveraging} use primitive shapes and smooth hinge losses to penalise collisions, whereas TrajOpt \cite{trajopt} applies approximate convex decomposition to leverage efficient convex shape distance computations. These methods are not suitable for real-world planning in unknown environments as prior knowledge about objects is not always available.

Sampling-based algorithms such as RRTs and PRMs require only binary occupancy information to represent the environment. This information can be stored in a simple 3D voxel grid whereby the environment is discretised into a regular grid, with each voxel storing the binary occupancy of that position in space. While this method provides fast memory access, it can require a large amount of memory to represent an environment. Octree methods provide a more memory-efficient representation with Octomap being a commonly used framework \cite{octomap}.

In contrast, trajectory optimisation-based motion planners require gradients. While stochastic algorithms, such as STOMP, obtain gradients using binary occupancy information, non-stochastic approaches require a continuously varying environment representation from which gradients can be obtained. CHOMP, TrajOpt, and GPMP2 use \acrfullpl{ESDF} to represent the environment \cite{chomp,trajopt,gpmp2}. \acrshortpl{ESDF} have been shown as an effective method for use in static environments however their CPU compute times generally prohibit real-time performance in a dynamic environment and are thus typically pre-computed and assumed to be static \cite{ITOMP}. Further, they commonly require the integration of an occupancy grid prior to computing an \acrshort{ESDF}. GPMP2 \cite{gpmp2} assumes this to be given, while other approaches can compute an \acrshort{ESDF} from alternative representations such as an Octomap.

To improve the computation time for \acrshort{ESDF} updates, Lau et al. presented an efficient method for incrementally updating an \acrshort{ESDF} from occupancy maps \cite{Lau2010}. Oleynikova et al. extended this approach with Voxblox to build and update \acrshortpl{ESDF} incrementally out of \acrshortpl{TSDF} \cite{voxblox}. Usenko et. al. introduced `ewok' which uses a fixed-size sliding window around the position of a \acrshort{MAV} to incrementally build \acrshortpl{ESDF} from occupancy \cite{Usenko2017}. Han et. al. use doubly-linked lists to present a time-efficient method of incrementally building \acrshortpl{USDF} \cite{Fiesta}. The aforementioned incremental methods have been demonstrated in 3D path planning environments for \acrshortpl{MAV}, however to our knowledge, they have not been demonstrated for use in whole-body motion planning for articulated systems or mobile manipulation platforms.


As the computations required for computing occupancy information and \acrshort{ESDF} are inherently SIMD-parallelisable, significant research has also been conducted into optimising signed distance field calculations on GPUs. J\"ulg et. al. present a comprehensive comparison of fast exact 3D \acrshort{EDT} implementations \cite{GPUVoxelsMobile}. In particular, they show the Parallel Banding Algorithm (PBA), developed by Cao et. al. \cite{PBA}, to be ``well suited for fast online GPU-based distance field computation". 

\subsection{Real-World Systems for Dynamic Obstacle Avoidance}
The vast majority of fully integrated systems have focused on mobile aerial robots. Voxblox integrated a local trajectory optimisation planner in static environments, while FIESTA uses kinodynamic path searching.

Alwala and Mukadam built upon the GPMP2 framework to present Joint Sampling and Trajectory Optimisation (JIST) and demonstrate effective collision avoidance in 2D navigation tasks as well as on a 7-DoF Sawyer robot arm \cite{Alwala2020} in simulation. However, JIST still uses pre-computed signed distance field calculations and was not verified on a real-world system with live sensor data.

Kaldestad et. al. \cite{Kaldestad2014} provide the first demonstration of collision avoidance in real-time on a real robot using parallel GPU processing. They calculate virtual forces to send to the robot impedance controller on a 7-DoF KUKA Arm. However, they use a restricted 2.5D environment model that is reset every time new depth sensor data is processed.

Hermann et. al. demonstrate mobile manipulation planning and re-planning on a GPU to operate in unknown environments by using grid-based planning techniques \cite{HermannMobileReplanning, GPUVoxels}. The planning times reported are an order of magnitude greater than those achieved in similar tasks using optimisation-based methods \cite{gpmp2}. The authors also observed that their method resulted in re-planning times that vary depending on how far along the current trajectory a new obstacle is observed. 
J\"ulg et al. \cite{GPUVoxelsMobile} built upon the GPU-Voxels framework \cite{GPUVoxels} to introduce highly optimised GPU calculations for \acrshortpl{EDT}. Due to the high speed of performing \acrshortpl{EDT}, they demonstrate real-time potential-field based motion planning of mobile aerial robot platforms in a fully 3D environment. Exploring the possibility of planning manipulator motions was beyond the scope of their work, motivating the work presented here.

To our knowledge, real-time receding-horizon trajectory optimisation has not been performed online in an unknown environment for a mobile manipulator. 


\section{Signed vs Unsigned Distance Fields}
Despite distance transforms being prominent in the literature for motion planning and \acrshort{MAV} path planning, we could not find any justification for whether distance fields should be signed or unsigned in motion planning.
For implementations such as wavefront planners, one would expect a signed distance field to have no impact in a static environment as long as the initial starting state is not in collision. In contrast, initial trajectories for a trajectory-optimisation based motion planner may start in collision (particularly when using the common ``straight-line" initialisation); these approaches require gradients to perform updates and it would seem intuitive to require a signed distance field and continuous gradients throughout an obstacle to provide gradients that `push' the trajectory out of collision. To verify this, we performed a series of experiments to analyse and compare trajectories generated using signed and unsigned distance fields.

\subsection{Methods}
Experiments were carried out in simulation for 2D path planning, 7-DoF arm manipulation (Franka Panda), and whole-body motion planning with a 5-DoF manipulator on a holonomic base (\acrshort{HSR}, 8-DoF). For each of the three cases, we generated 18 different robot states to produce a set of 153 pairs of start and goal configurations. We generated an obstacle set comprising of 100 cuboids of randomly generated size in the range of \SIrange{0.0}{1.0}{\metre} for each dimension; each obstacle was associated with a randomly generated position in the workspace. With both \acrshortpl{USDF} and \acrshortpl{SDF}, we used GPMP2 with consistent parameter settings as a motion planner for each of the $15\,300$ motion planning problems. We used the Levenberg-Marquardt method of optimisation with the initial damping parameter set as $0.01$. The optimisation stopped if there was a relative decrease in error smaller than $10^{-5}$ or if it reached $100$ iterations.

\subsection{Results}
To calculate the `failure rate', we exclude cases in which the planner failed to find a collision-free trajectory using either the \acrshort{SDF} or \acrshort{USDF}. Similarly, we only compare trajectory costs across collision-free trajectories since trajectories that result in a collision would not be executed. 

A summary of our findings is presented in Table \ref{table:sdf_comparison_table}. In the \textit{2D Navigation} experiments, we find that our intuition of \acrshort{SDF} gradients `pushing' the trajectory out of collision is confirmed, with plans that use USDFs having a failure rate that is 18.3 times greater. For a 2D planning example, this can be explained by entire query states being inside obstacles with no distance or distance-gradient information and the only gradients to `pull' out obstacles come from the Gaussian Process smoothness factors. However, in higher dimensions we found that \acrshortpl{SDF} provide no practical advantage over \acrshortpl{USDF} for motion planning tasks in absolute terms, with almost all planning problems being solved successfully in both cases. We further validated this finding in complex narrow passage examples on a 7-DoF Panda arm manipulator where \acrshortpl{SDF} and \acrshortpl{USDF} performed equally well from infeasible straight line initialisations through the obstacle. This can be explained in articulated systems, such as in the \textit{Arm} and \textit{Whole-Body} tasks, by gradients being available at multiple other locations that are not in collision; these gradients will assist in `pulling' states out of collision.

Considering that a \acrshort{SDF} calculation typically takes around twice as long to compute as a \acrshort{USDF}, we conclude that \acrshortpl{USDF} are preferable to use on articulated and whole-body systems, particularly when operating in dynamic environments since the faster re-planning speed enables better adaption to changing environments. 


\begin{table*}[t]
\caption{Comparison between using SDFs and USDFs for collision avoidance}
\label{table:sdf_comparison_table}
\begin{tabularx}{\linewidth}{| c | C | C | C | C |}
\hline
& & USDF & SDF & Relative Difference (USDF/SDF) \\
\hline
\multirow{3}{*}{Failure Rate (\%)} 
& Navigation & $2.9$ & $\mathbf{0.16}$ & $18.3$\\ 
& Arm & $\mathbf{0.00028}$ & $0.00037$ & $0.75$ \\ 
& Whole-Body & $0.00092$ & $\mathbf{0.00018}$ & $5.0$\\
\hline
\multirow{3}{*}{Iterations ($\mu \pm \sigma$) } 
& Navigation & $\mathbf{7.57 \pm 5.90}$ & $8.24 \pm 6.66$  & $0.9$ \\ 
& Arm & $\mathbf{7.03 \pm 7.34} $ & $7.12 \pm 7.71$ & $1.0$ \\ 
& Whole-Body & $5.84 \pm 4.96$ & $\mathbf{5.81 \pm 4.91}$ & $1.0$ \\
\hline
\multirow{3}{*}{Valid Trajectory Cost ($\mu \pm \sigma$)}
& Navigation & $68.6 \pm 504$ & $\mathbf{61.5 \pm 379}$ & $1.1$ \\
& Arm & $\mathbf{7.82 \pm 60.7}$ & $7.83 \pm 60.73$ & $1.0$ \\
& Whole-Body & $2.81 \pm 14.7$ & $\mathbf{2.74 \pm 11.5}$ & $1.0$ \\
\hline
\end{tabularx}
\end{table*}

\section{Real-time Scene Reconstruction}
%
We used the state-of-the-art mapping packages Voxblox \cite{voxblox} (CPU), FIESTA \cite{Fiesta} (CPU), and GPU-Voxels \cite{GPUVoxels} (GPU) to generate distance fields from live sensor data and provide direct query access to the obstacle factors used in GPMP2. Voxblox generates \acrshortpl{SDF} whereas GPU-Voxels and FIESTA produce \acrshortpl{USDF}. To provide a more thorough comparison, we adapted GPU-Voxels and FIESTA to optionally produce \acrshortpl{SDF} using their native methodology. We achieve this by calculating distance fields for both the occupancy map and the inverse occupancy map; the values are then subtracted to produce a \acrshort{SDF}. In the case of GPU-Voxels, this was done such that the inverse distance field could be calculated in parallel. In practice, however, GPUs are still limited by the number of threads available and so for large environments, the two distance transforms will still be performed sequentially. For FIESTA, we implemented the inverse distance transform in a similar manner to how the original distance transform is calculated but reversed the roles of unoccupied and occupied cells found in ray-casting. 

To compare the performance of the mapping frameworks, we used the Cow and Lady dataset, as first presented by Oleynikova et. al. \cite{voxblox}. We chose this dataset because it uses a small, indoor scene with multiple objects and accessories in the room; similar to a typical environment in which a mobile service robot might be deployed. The dataset features a rosbag file in which real RGB pointcloud data was collected using a Kinect v1 depth camera, along with published ground-truth pose transforms of the camera frame using a vicon sensor.

We evaluated each package across a range of resolutions, while retaining full spatial coverage, on the \SI{142}{\second} dataset. For each evaluation, the log file was played in real-time to simulate live operation and the update rate was recorded using an 8-core Intel Core i7-9700 CPU @ \SI{4.50}{\giga\hertz} and \SI{2133}{\mega\hertz} DDR4 RAM. We found that FIESTA did not operate successfully using multiple threads and so ran this package with a single thread; Voxblox was run in multi-threaded mode with 8 threads. GPU-Voxels was run on a Nvidia RTX 2060 GPU (1920 CUDA cores).

\subsection{Results}

\begin{table*}[t]
\caption{Comparison of time taken (\si{\milli\second}) to compute distance fields from point cloud data in state-of-the-art mapping frameworks}
\label{table:benchmarking_table}
\begin{tabularx}{\linewidth}{| c | C | C | C | C |}
\multicolumn{2}{ c }{}&\multicolumn{3}{ c }{} \\ \cline{3-5}  
\multicolumn{2}{ c |}{}&\multicolumn{3}{ c |}{Resolution (\si{\metre})}\\
\hline
                            & Distance Field & 0.025 & 0.05 & 0.10\\
\hline
\multirow{2}{*}{Voxblox (CPU)} 
& Signed & $1232.5 \pm 677.6$ & $210.1 \pm 124.1$ & $41.4 \pm 29.4$
\\ & Unsigned & - & - & - \\
\hline
\multirow{2}{*}{Fiesta (CPU)} 
& Signed & $1652.0 \pm 526.1$ & $176.8 \pm 26.9$ & $178.1 \pm 25.6$
\\ & Unsigned & $176.4 \pm 206.5$ & $31.4 \pm 17.7$ & $11.6 \pm 2.7$ \\
\hline
\multirow{2}{*}{GPU-Voxels (GPU)} 
& Signed & $\mathbf{36.2 \pm 8.3}$ & $\mathbf{17.5 \pm 0.4}$ & $\mathbf{6.9 \pm 1.6}$ \\
& Unsigned & $\mathbf{25.3 \pm 5.6}$ & $\mathbf{13.4 \pm 2.9}$ & $\mathbf{5.6 \pm 1.4}$ \\
\hline
\end{tabularx}
\end{table*}

Results of our mapping framework comparison are shown in Table \ref{table:benchmarking_table}. For Voxblox, we present the results for a propagation distance of \SI{0.8}{\metre}; we require a propagation distance at least as large as the sum of the maximum distance penalised in trajectory optimisation and the maximum radius of the spheres used to represent a robot's collision model. The maximum sphere radius used in our collision model of the \acrshort{HSR} is \SI{0.3}{\metre} and the maximum penalty distance used in our later experiments is \SI{0.5}{\metre}. We note that at the smallest propagation distance for which we measured performance, \SI{0.1}{\metre}, FIESTA and GPU-Voxels still outperformed Voxblox.

\begin{figure}[t]
  \centering
    \includegraphics[width=1.0\columnwidth]{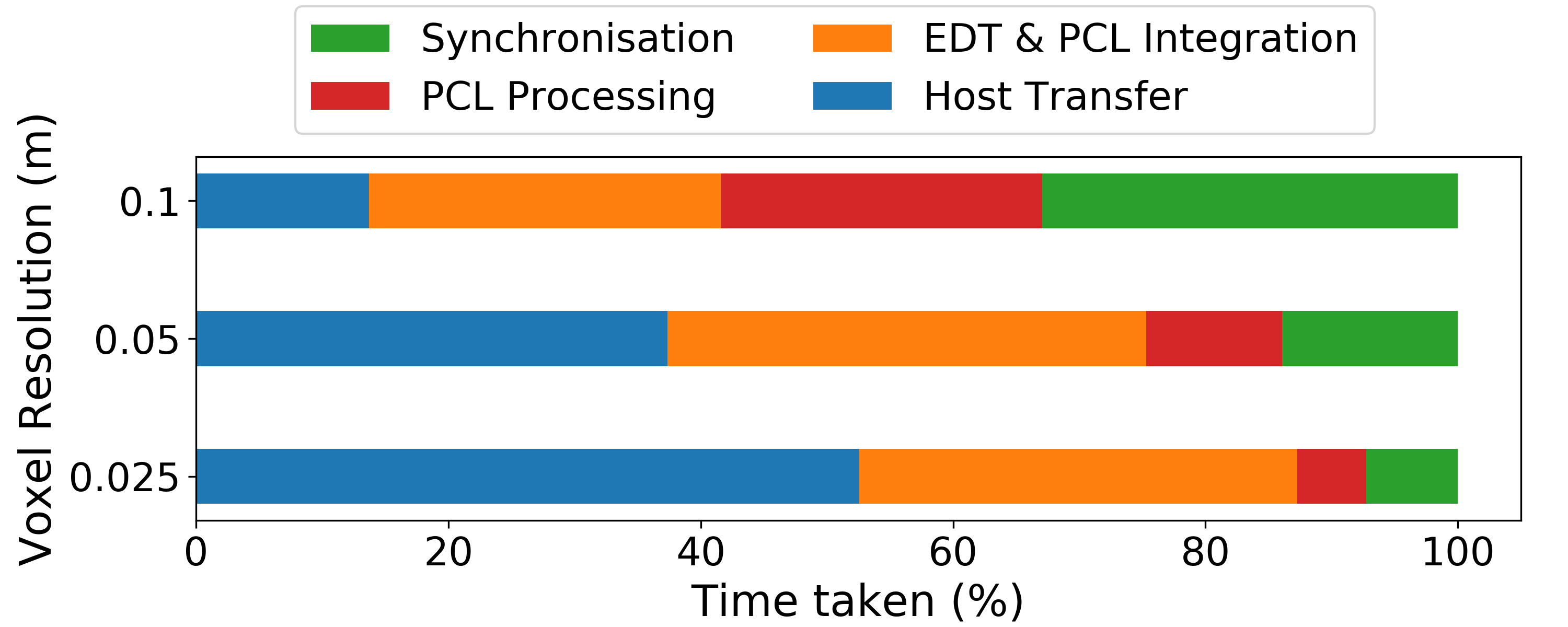}
    \includegraphics[width=1.0\columnwidth]{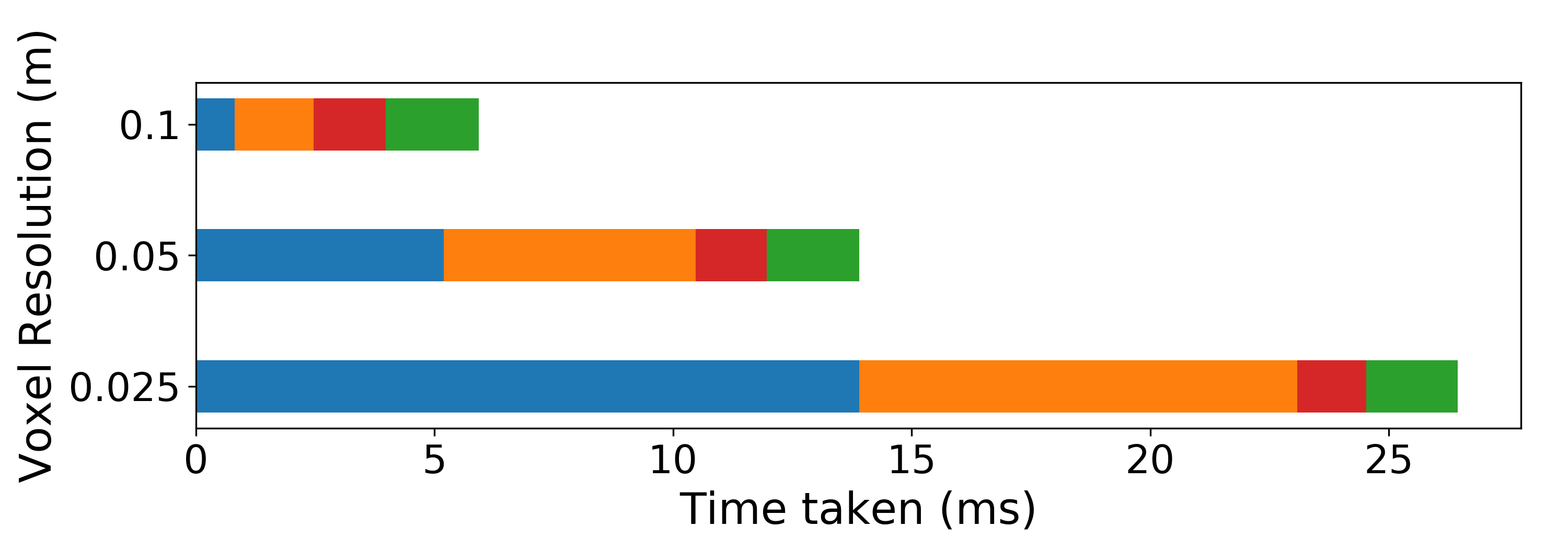}
  \caption{A breakdown of the time spent in our GPU-Voxels-based update loop. \textit{Synchronisation} -- the time spent finding camera pose transforms within a time tolerance of pointcloud timestamps in the update queue. \textit{PCL Processing} -- time taken to resize and apply reference frame transformations to the incoming pointclouds prior to integration. \textit{EDT and PCL Integration} -- pointclouds are integrated into a probabilistic voxelmap via raycasting and a \acrshort{EDT} is then calculated using the PBA algorithm. \textit{Host Transfer} -- time spent copying the full \acrshort{EDT} from the GPU to memory on the host computer.}
  \label{fig:gpu_voxels_breakdown}
  \vspace{-10pt}
\end{figure}

Our results show that despite the additional time required by GPU-Voxels to transfer the distance field from the GPU, it is the fastest mapping framework in all cases. Figure \ref{fig:gpu_voxels_breakdown} illustrates a breakdown of the time spent in our GPU-Voxels update loop. While the time spent in \textit{Synchronisation} and \textit{PCL Processing} is essentially constant, we see that as we increase the voxel resolution and grid size, the time taken to transfer data from the GPU to the host becomes more significant.

\section{Integrating Mapping and Motion Planning}
As previously discussed, we elected to use GPMP2 as our motion planner of choice. Using the GPMP2 framework, collision avoidance is implemented via the use of regularly spaced, time-indexed obstacle factors in a factor graph. Obstacle factors query the distance field to allocate a hinge-loss obstacle cost and an associated gradient for optimisation. In this work, all obstacle factors were linked to use the real-time updated distance field as maintained by the package used. Based on the results shown in the previous section, GPU-Voxels was used as the mapping framework for the rest of this work.  

We run the mapping concurrently on a separate thread from the motion planning framework. Fast distance query access is provided to the obstacle factors by running GPU-Voxels within the same ROS node and providing them with pointers to the \acrshort{EDT} memory address. 

While we experimented with different methods of maintaining and updating a factor graph, we found the most effective way of planning to be that shown in Algorithm \ref{algo:1}. After providing the algorithm with a goal pose, consisting of a base pose ($x, y, \theta$) and joint configuration for the arm, we estimate the time needed to achieve the goal state and construct a factor graph accordingly with a fixed $\delta_{t}$ between variable nodes. Prior factors are imposed on the current robot state and the goal state. We use a straight-line trajectory initialisation and optimise until convergence criteria are met (relative error decrease of $0.01$ or $50$ iterations). The trajectory is then interpolated to a finer discretisation and executed on the robot.

After the initial trajectory execution, we use timer callbacks to regularly check whether the current trajectory is collision-free and within an error tolerance of the cost when it was first calculated; this is achieved using the real-time updated \acrshort{EDT}. If either of these criteria are not met, we obtain our current pose, re-estimate how long it will take to achieve the goal position, re-build the graph and re-optimise. To re-optimise, we use the previous trajectory and re-fit it to the new graph discretisation. This enables us to retain information from previous optimisations and we found it to provide $\approx30\%$ speed-up in optimising when compared to using a straight-line initialisation each re-planning iteration. 

While re-building the graph each iteration may seem unnecessary, it provides multiple benefits. Firstly, this method avoids book-keeping and pruning of factors in the past. Secondly, it affords us the freedom to easily re-parametrise and change the planning horizon as new information is acquired.
\begin{algorithm}
	\caption{Motion Planning Pipeline}
	\begin{algorithmic}[1]
		\renewcommand{\algorithmicrequire}{\textbf{Input:}}
		\renewcommand{\algorithmicensure}{\textbf{Usage:}}
		\REQUIRE Goal state $\mathbf{x}_g$
		\ENSURE  Re-planning and execution
		\\ \textit{Initial optimisation} :
		\STATE $\mathbf{x}_c = \text{getCurrentPose}()$
		\STATE $\mathbfcal{P} = \text{estimateParameters}(\mathbf{x}_c, \mathbf{x}_g)$ 
		\STATE $\mathbfcal{G} = \text{buildGraph}(\mathbf{x}_c, \mathbfcal{P})$
		\STATE $\tau_{straight} = \text{initialiseTrajectory}(\mathbf{x}_c, \mathbf{x}_g,  \mathbfcal{P})$
        \STATE $\tau  =\text{optimise}(\tau_{straight}, \mathbfcal{G})$
		\STATE $\tau_{int} = \text{interpolateTrajectory}(\tau )$
		\STATE $\text{executeTrajectory}(\tau_{int})$				\\ \textit{Re-planning} :
        \WHILE{$!\text{isGoalReached}()$}
        
        \IF{stillValid($\tau$)}
            \STATE Continue
	    \ENDIF

		\STATE $\mathbf{x}_c = \text{getCurrentPose}()$
		\STATE $\mathbfcal{P} = \text{estimateParameters}(\mathbf{x}_c, \mathbf{x}_g)$ 
		\STATE $\mathbfcal{G} = \text{buildGraph}(\mathbf{x}_c, \mathbfcal{P})$
		\STATE $\tau_{init}  = \text{refitTrajectory}(\mathbf{x}_c, \mathbfcal{P})$
		\STATE $\tau  = \text{optimise}(\tau_{init}, \mathbfcal{G})$
		\STATE $\tau_{int} = \text{interpolateTrajectory}(\tau )$
		\STATE executeTrajectory($\tau_{int}$)
		\ENDWHILE
	\end{algorithmic}
	\label{algo:1}
\end{algorithm}

\begin{figure*}[htbp]
    \centering
    \subfloat[][\label{fig:sim_floor_no_obs}]{%
       \includegraphics[width=1.0\columnwidth,trim={0cm 1.5cm 0cm 0cm},clip]{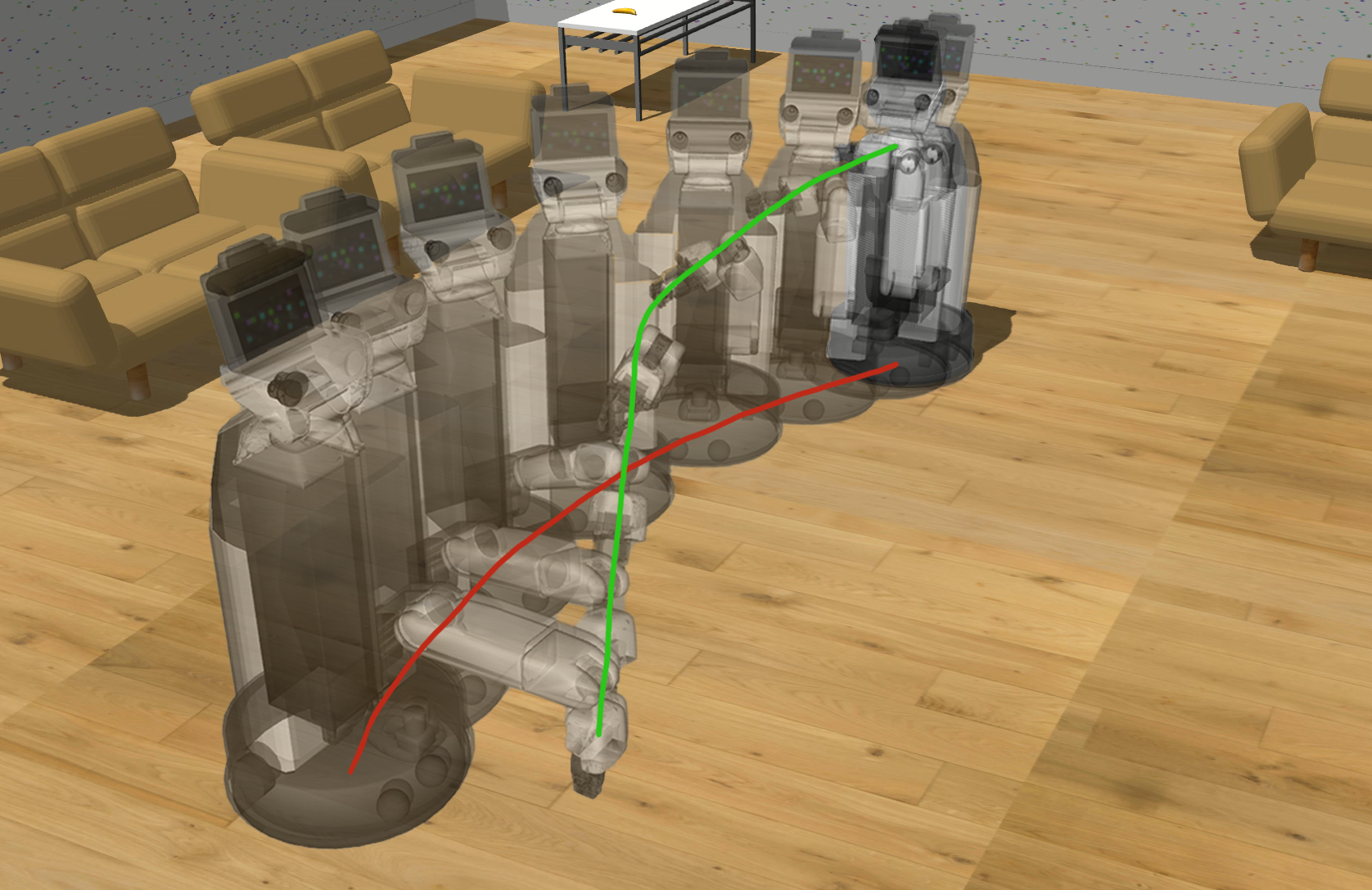}}
    \hfill
    \subfloat[][\label{fig:sim_floor_obs}]{%
        \includegraphics[width=1.0\columnwidth,trim={0cm 1.5cm 0cm 0cm},clip]{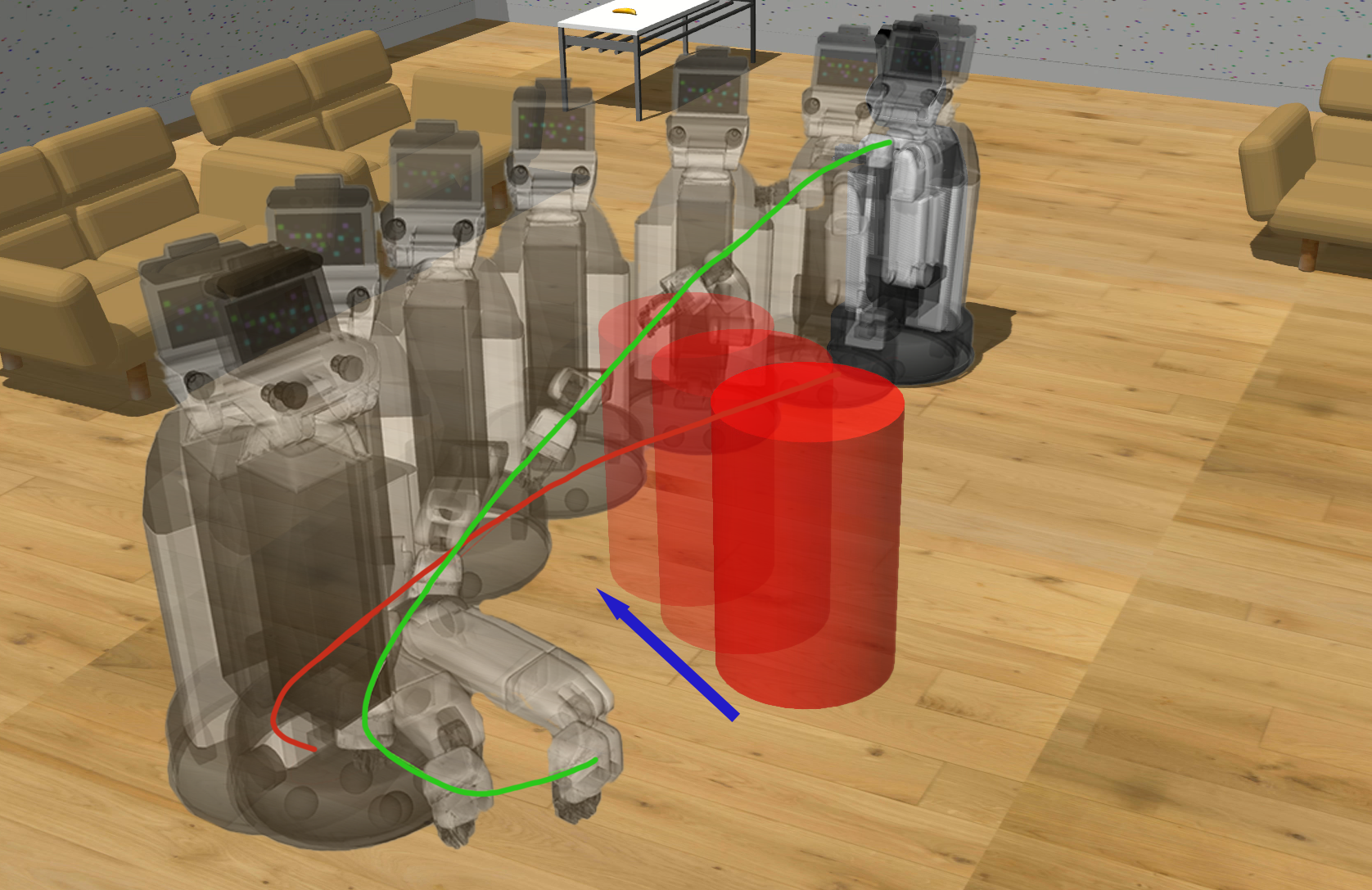}}
  \caption{Simulated task in which the robot is required to achieved a `pickup' goal state. During execution, a large (red) obstacle traverses across the planned robot trajectory, forcing it to re-plan online and adapt to a collision-free trajectory. Green lines illustrate the planned end-effector trajectory while red lines show the planned path of the base.}
  \label{fig:simulation_pickup}
\end{figure*}

\subsection{Simulation Experiments} 
We conducted simulation experiments using an 8-core Intel Core i7-9700 CPU @ \SI{4.50}{\giga\hertz} and \SI{2133}{\mega\hertz} RAM. GPU calculations were performed on a NVIDIA GeForce RTX 2060 graphics card (1920 CUDA cores).

The implementation described in the previous section was tested on a range of whole-body motion tasks in the presence of moving obstacles, from general navigation tasks within a room to reaching in shelves and picking objects up from the floor. Figure \ref{fig:simulation_pickup} illustrates a scenario in which the robot was tasked with achieving a goal state from which it can pick up an object from the floor. During execution, a large object (red cylinder) moves into the planned path of the robot, forcing it to re-plan smoothly around the obstacle.

We found that the robot robustly avoids moving obstacles, however this is still dependent on whether the robot perceives the object---if the head camera cannot see an obstacle then inevitably it will not be registered in the distance field used for motion planning.

While we can evaluate whether our current trajectory is collision-free and within error tolerances at \SI{250}{\hertz}, when a new trajectory is requested, our re-plan loop runs at \SI{10}{\hertz}.

\begin{figure*}[htbp!]
    \centering
    \subfloat[][\label{fig:real_start}]{%
       \includegraphics[width=0.495\linewidth,trim={0cm 4cm 0cm 0cm},clip]{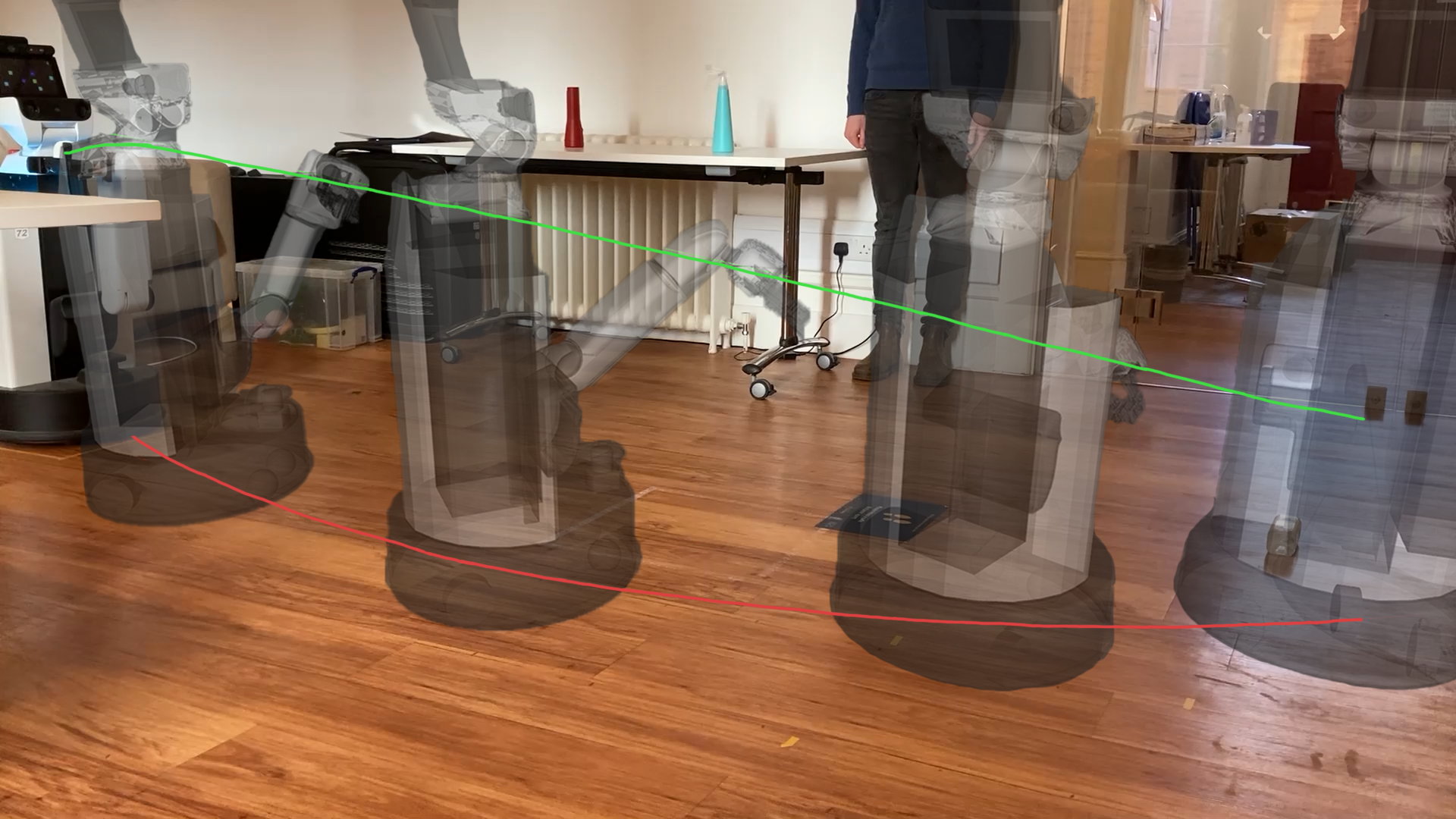}}
    \hfill
    \subfloat[][\label{fig:real_mid}]{%
        \includegraphics[width=0.495\linewidth,trim={0cm 4cm 0cm 0cm},clip]{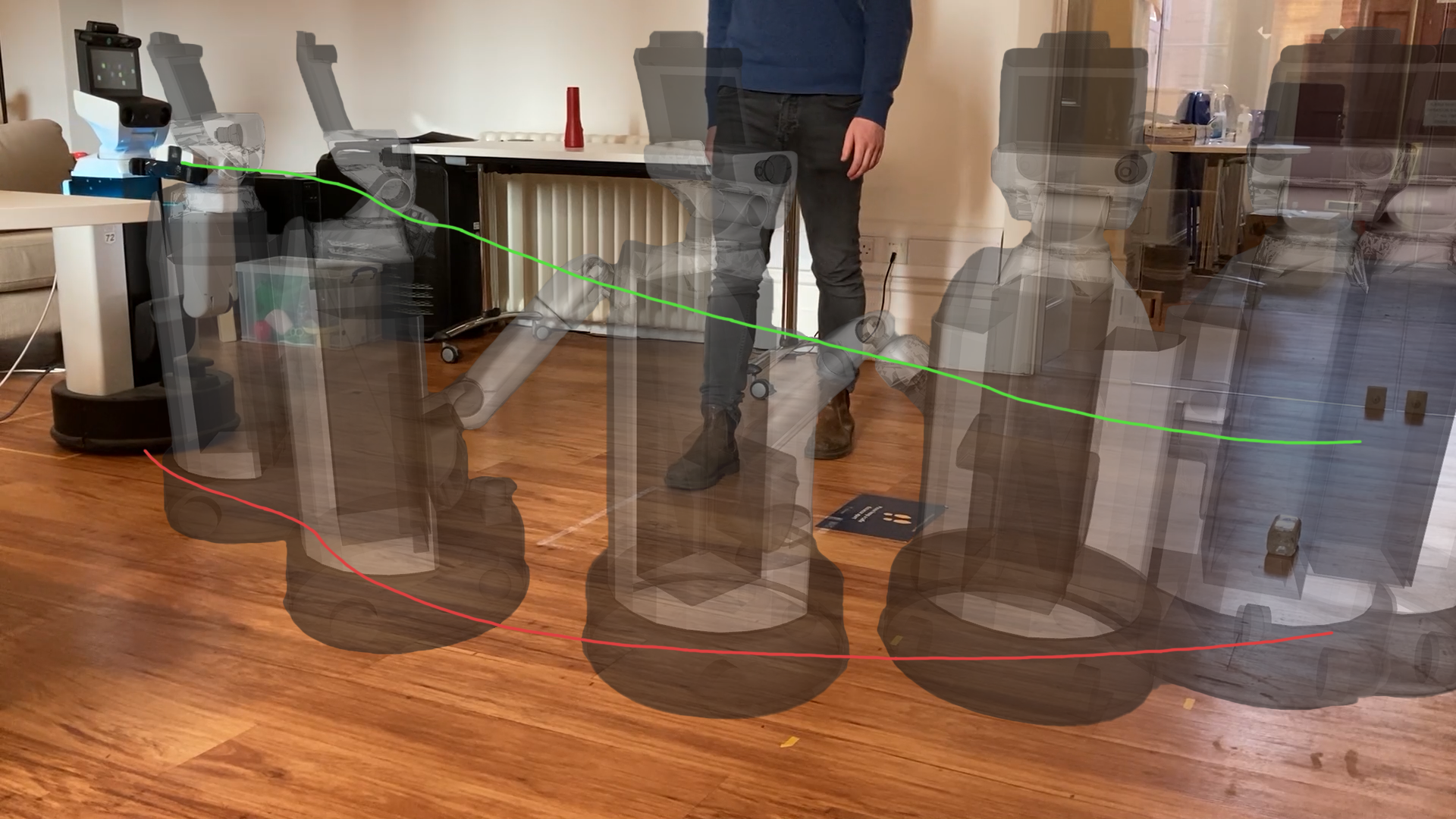}}
    \vfill
    \subfloat[][\label{fig:real_start_voxels}]{%
       \includegraphics[width=0.495\linewidth,trim={0cm 4cm 0cm 0cm},clip]{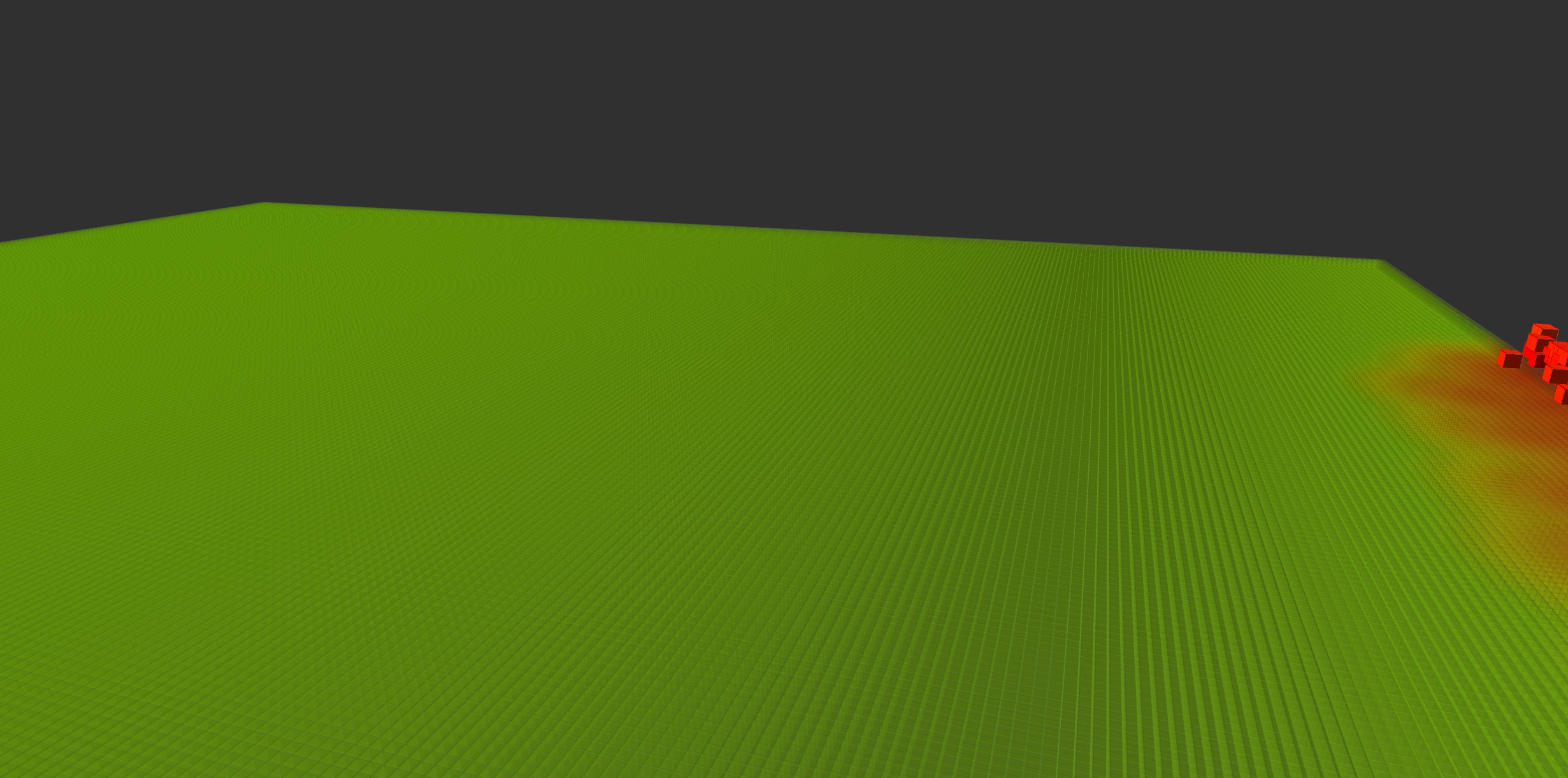}}
    \hfill
    \subfloat[][\label{fig:real_mid_voxels}]{%
        \includegraphics[width=0.495\linewidth,trim={0cm 4cm 0cm 0cm},clip]{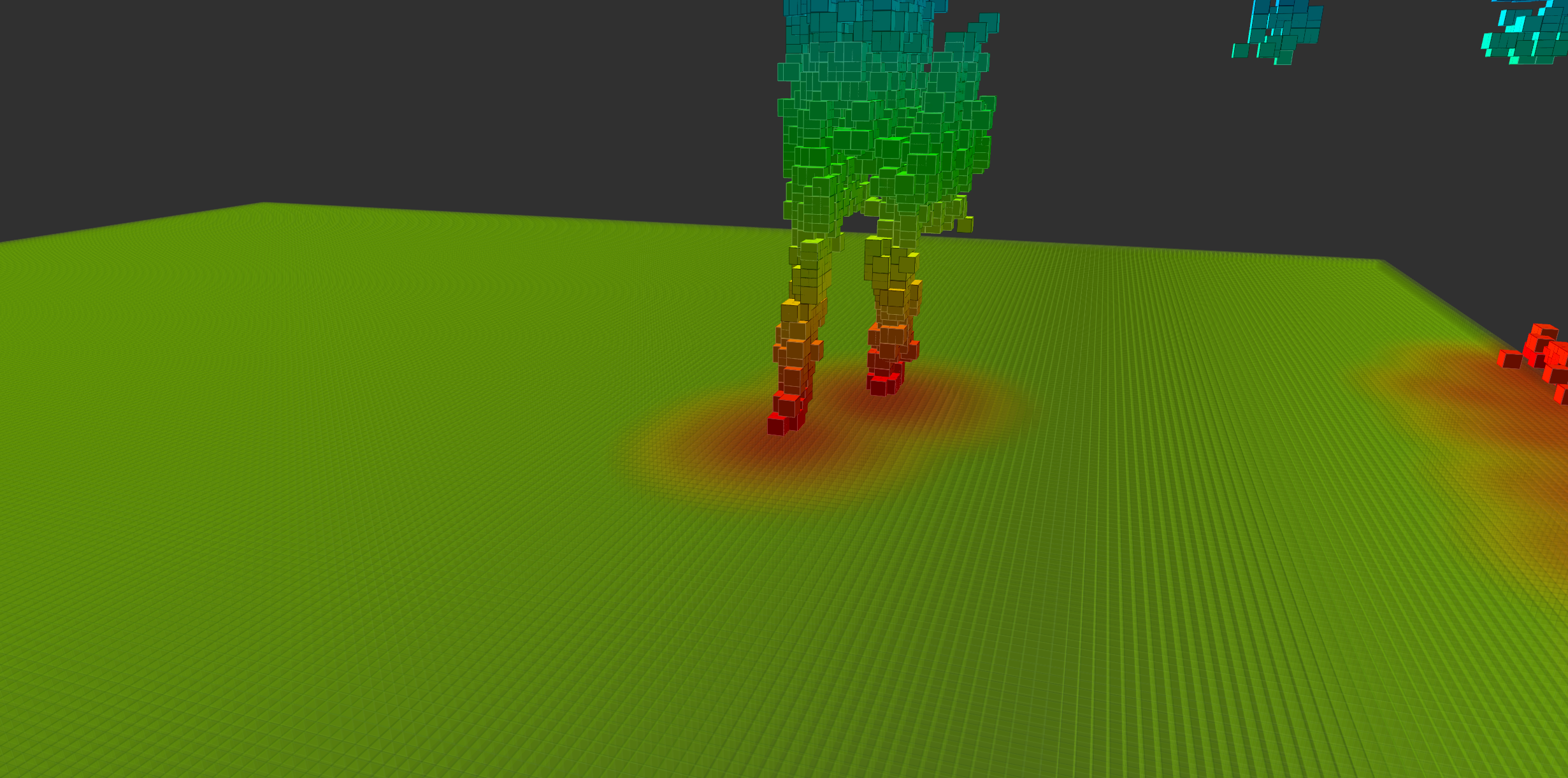}}
  \caption{Demonstration of re-planning in an unknown, dynamic environment on a physical \acrshort{HSR}. The robot was tasked with picking up a bottle from the floor. During execution, a human demonstrator walked towards the planned robot trajectory, requiring a re-planned trajectory to be calculated and executed to avoid collision. Green lines illustrate the planned end-effector trajectory while red lines show the planned path of the base. The bottom row illustrates the distance field generated at the corresponding times; we show both a plane through the distance field and a pointcloud after thresholding at zero distance. The interactive re-planning can more readily be appreciated in the accompanying video.}
  \label{fig:real_robot}
\end{figure*}

\subsection{Hardware Experiments} 
We implemented our system for execution on a \acrshort{HSR}. The robot has an onboard 4th Gen Intel Core i7 (\SI{16}{\giga\byte} RAM) processor on which joint controllers and sensing operates. The RGB-D sensor is an Xtion PRO LIVE delivering pointclouds at VGA resolution similar to the sensor in the benchmark dataset. Motion planning and mapping  were performed on an external laptop with an Intel Core i7-10875H CPU, \SI{32}{\giga\byte} \SI{2666}{\mega\hertz} RAM, and a NVIDIA GeForce RTX 2070 SUPER GPU (2560 CUDA cores). To increase throughput of point-cloud processing and retain \SI{30}{\hertz} pointcloud updates, we streamed depth images to the laptop via a wired connection and performed point-cloud conversion locally. To provide a good resolution for manipulation tasks in confined environments such as shelves, we used a voxel resolution of \SI{2.5}{\centi\metre} and a grid size of $320 \times 320 \times 128$.
We tested our implementation in a variety of environments with moving obstacles:
\subsubsection{Shelf Pickup}
The robot is required to reach deep into a shelf to pick up an object; during execution a human demonstrator walks into the planned robot trajectory. Later on in the same task, a wooden plank is placed across the end-effector path. 

\subsubsection{Floor Pickup}
The robot is required to travel to a location and pick up a bottle from the floor. During execution, a human demonstrator walks into the planned robot trajectory. 

\subsubsection{Table Pickup}
The robot is tasked with picking up an object from a table. During execution, a human demonstrator moves an object across the planned trajectory.

In all three cases, the robot re-planned and successfully achieved the desired goal state without collisions. Figure \ref{fig:hardware_timelapse} shows the \textit{shelf pickup} and how the robot's resultant trajectory takes a different path to its initial plan in order to avoid an obstacle that moves into the scene during execution. Figure \ref{fig:real_robot} shows the environment perception and re-planning during the \textit{floor pickup} task. In particular, Figure \ref{fig:real_mid_voxels} emphasises the clarity of the distance field reconstruction obtained in real-time. The real-time update of the \acrshort{SDF} can also be seen in the accompanying video.

\section{Discussion and Future Work}
For mapping, in contrast to the incremental methods discussed, a limitation of GPU-Voxels is its fixed-size memory allocation on the GPU. In the current release of GPU-Voxels, the maximum number of blocks is limited to $65\,535$. With $1024$ threads per block, this corresponds to a maximum environment grid of $67\,107\,840$ voxels that can be computed in a single invocation. The maximum grid size is thus restricted to approximately $704 \times 704 \times 128$, or $\SI{17.6}{\metre} \times \SI{17.6}{\metre} \times \SI{3.2}{\metre}$ at a \SI{2.5}{\centi\metre} resolution. While this may be prohibitive for large scale, multi-room operation at high resolution, for robots operating within a single workspace however, this is likely sufficient to provide high resolution mapping at a superior compute speed. In practice, one is not likely to need a high resolution for planning across large distances and a coarse representation could be used while maintaining a finer resolution in the vicinity of the robot. Current CUDA devices can support much larger number of blocks than the GPU-Voxels limit and so we believe that with further work, this package could support larger environment grid sizes.

As mentioned previously, iSAM-based re-planning can be used to provide re-planned trajectories much faster than planning from scratch --- prior work \cite{gpmp2} cites a possible one order of magnitude speed-up depending on task. The effectiveness of this method is particularly applicable when tail factors are changed, such as the goal state prior. This is relevant when tracking a moving object to grasp and is a promising feature which we hope to include in future work with the integration of object detection.

In building a practical system which performed both motion planning and real-time mapping of the environment, we make two notable observations which require further work. Firstly, the positioning of the head camera quickly became the primary limitation for the system. As the robot is executing and re-planning a trajectory, determining where to aim the head camera---i.e. Next Best View planning for receding-horizon trajectory optimisation---is an interesting problem in itself which is more commonly explored in the context of building 3D models of objects or structures \cite{Collander2021, Border}, rather than for use in motion planning. Na\"ive heuristics, such as always looking in the direction of motion or looking a specified distance along the planned trajectory, are prone to failure cases, in particular on curved trajectories. We mention this problem as an interesting observation for the community and as a project for further work. The second limitation is that inherent with performing trajectory optimisation which assumes a static environment. In previous work, we highlighted that using trajectory optimisation for re-planning in the presence of moving obstacles can result in trajectories that repeatedly plan to go into the path of the moving obstacle. In practice, this leads to sub-optimal trajectories. In future work, we will integrate the work presented here with methods such as predicted composite signed distance fields to account for the predicted future motion of moving obstacles as in time-configuration space planning \cite{PredictedSDFs}.


Due to the common use of factor graphs in state estimation, another avenue for future work is to interleave planning using GPMP2 with state estimation. To this end, Mukadam et. al. demonstrated simultaneous trajectory estimation and planning (STEAP) of a PR2 robot operating within a known 3D workspace \cite{STEAP}. In their work, they repeatedly perform inference on a factor graph spanning from the start state to the goal state, while adding measurement factors, to incorporate new sensor measurements and observations, during execution. The resultant trajectory after each optimisation provides a solution to both the motion planning and state estimation problems. The primary limitation they cite is that it can only operate in ``known, static environments" because the \acrshort{SDF} computation provides a ``major, computational bottleneck". We believe that by leveraging the contributions presented in our work, STEAP could be extended to provide simultaneous mapping, localisation and planning in dynamic environments.

A key implementation note is that point cloud observations are useful not only for the occupied space but also for clearing space via ray-casting. In our work, we did not wish to classify the floor as a collision object in order to provide stronger gradients around real obstacles. Points registered within \SI{3}{\centi\metre} of the ground were used in the ray-casting for clearing, however the occupied point at the end of the ray was not inserted into the voxelmap.

Finally, transferring information between CPU and GPU (device-to-host) became a more dominating factor in computing \acrshortpl{EDT} for finer resolutions. To avoid this requirement, motion planning directly on the GPU could be explored. Previous work based on GPU-Voxels performed grid-based motion planning and environment mapping directly on the GPU to eliminate the transfer costs \cite{GPUVoxels, GPUVoxelsMobile, HermannMobileReplanning}.
Hence, a possible avenue for future work could look into implementing optimisation-based motion planning on the GPU.

\section{Conclusion}
This paper presents a fully integrated system to update the environment representation in real-time and perform re-planning in dynamic environments. We show experimentally that signed distance fields provide no significant benefit to motion planning on articulated systems when compared to their cheaper-to-compute unsigned counterparts. We analyse a selection of state-of-the-art mapping libraries and show GPU-Voxels to be a superior candidate for using in whole-body trajectory optimisation problems. We integrate GPMP2 with GPU-Voxels to produce a hybrid mapping and motion planning system which can provide real-time mapping and whole-body motion planning to smoothly avoid moving obstacles. Our findings are verified both in simulation and on a physical \acrshort{HSR} across a range of tasks and successfully re-plan safely in response to dynamic obstacles.

\bibliographystyle{IEEEtran}
\bibliography{main}

\end{document}